%% file: main.tex
\documentclass[letterpaper]{article} 
\usepackage{aaai2026}  
\usepackage{times}  
\usepackage{helvet}  
\usepackage{courier}  
\usepackage[hyphens]{url}  
\usepackage{graphicx} 
\usepackage{natbib}  
\usepackage{caption} 
\usepackage{algorithm}
\usepackage{algorithmic}

\usepackage{newfloat}
\usepackage{listings}
\DeclareCaptionStyle{ruled}{labelfont=normalfont,labelsep=colon,strut=off} 
\floatstyle{ruled}
\newfloat{listing}{tb}{lst}{}
\floatname{listing}{Listing}
\title{Tracking the Temporal Dynamics of News Coverage of Catastrophic and Violent Events}
\author {
    Emily N. Lugos \textsuperscript{\rm 1},
    Maur\'{i}cio Gruppi \textsuperscript{\rm 1}
}
\affiliations {
    \textsuperscript{\rm 1}Department of Computing Sciences, Villanova University, Villanova, PA, USA\\
    elugos@villanova.edu, mgouveag@villanova.edu
}

\usepackage{bibentry}

\usepackage{enumitem}
\usepackage{multirow}
\usepackage{subcaption}

\begin{document}

\maketitle

\begin{abstract}
The modern news cycle has been fundamentally reshaped by the rapid exchange of information online. As a result, media framing shifts dynamically as new information, political responses, and social reactions emerge. Understanding how these narratives form, propagate, and evolve is essential for interpreting public discourse during moments of crisis. In this study, we examine the temporal and semantic dynamics of reporting for violent and catastrophic events using a large-scale corpus of 126,602 news articles collected from online publishers. We quantify narrative change through publication volume, semantic drift, semantic dispersion, and term relevance. Our results show that sudden events of impact exhibit structured and predictable news-cycle patterns characterized by rapid surges in coverage, early semantic drift, and gradual declines toward the baseline. In addition, our results indicate the terms that are driving the temporal patterns.
\end{abstract}


\section{Introduction}

The modern news cycle has undergone a profound transformation with the rise of digital publishing platforms and social media. News production is no longer constrained by daily print deadlines or broadcast schedules; instead, information is continuously published and instantaneously amplified across a global media ecosystem. During crisis events and high-impact incidents, such as natural disasters, mass shootings, terrorist attacks, and political upheavals, the acceleration of digital publishing generates an overwhelming volume of articles, updates, commentary, and secondary reporting \cite{livingston2003gatekeeping, nielsen2016challenges}. 

While this abundance of information increases access and visibility, it also introduces substantial noise, redundancy, and misinformation, making it increasingly difficult to distinguish reliable reporting from speculative or low-quality content.  
\citet{silva2021news} analyzed framing in \textit{The New York Times} (2000–2016), finding dynamic shifts in how mass shootings were presented, from gun access to mental illness or terrorism, illustrating the volatile nature of modern news cycles.
News coverage surges rapidly following an event’s onset, peaks as new developments emerge, and then gradually decays as public attention shifts elsewhere. Within this compressed attention window, the framing evolves, interpretations shift, and media framing changes as additional information, political responses, and social reactions emerge. Understanding how these narratives form, evolve, and decay is essential for journalists, policymakers, emergency responders, and researchers seeking to interpret public discourse during moments of crisis.

Semantic analysis has been previously used to track framing evolution in news reporting. \citet{park2025applying} used semantic network analysis to track how editorial narratives during COVID‑19 shifted thematically, from public health to economic concerns, illustrating measurable semantic drift in crisis reporting. \citet{geiss2025inflation} traced the inflation in crisis‑label usage from 1785 to 2020 by the U.K. newspaper \textit{The Times}, attributing rising narrative waves to political mediatization and attention steering, underscoring the increasing volatility of crisis narratives. \citet{an2017gets} described the patterns of media attention evolution as exhibiting a rapid rise and gradual decline.
Moreover, studies have found that the news coverage of violent and catastrophic events lead to public trust disruption, creation of myths, and psychological and mental-health impacts \cite{wormwood2019psychological, tierney2006metaphors}. 

In this study, we examine the coverage and semantic dynamics of destructive events throughout the phases of the online news publishers. We construct a large-scale corpus of 126,602 articles, related to 12 events, collected from online news sources. Events are selected from one of two categories: environmental disasters and acts of violence. Using a combination of keyword matching and embedding similarity, we identify documents whose content relate to the selected events. From the resulting corpus, we analyze the duration of an event’s coverage in media and model how the framing evolves from initial onset through peak coverage and into decline.
We then analyze how media framing shifts over time by tracking semantic sentiment, thematic focus, and framing patterns across multiple temporal phases: the breaking-news stage, the saturation phase, and the post-event stabilization period extending up to 30 days after event onset. This allows us to quantify framing drift and identify the linguistic and semantic drivers that shape public understanding as an event unfolds.

We present the following research questions:

\begin{description}[leftmargin=0pt]
    \item[\textbf{RQ1:}]  How long does an impactful event remain active in news cycles, and how does framing shift in time, as measured by sustained media coverage and semantic coherence?
    \item[\textbf{RQ2:}]  What linguistic factors are associated with the framing drift in event reporting?
\end{description}

Our findings show that temporal patterns in publication volume, framing drift, and dispersion mirror the dynamics of disaster and violent events, characterized by a rapid surge in media coverage following event onset and a gradual decline as attention diminishes. The former showing anticipation and preparedness before the event onset, followed by a broad discussion about relief, damage and recovery. The latter showing a sudden spike in volume, followed by a quick decline and more narrowly dispersed discussion. Our results also highlight the words driving the changes in news dynamics.

\section{Data}

We defined two major categories of events: (i) disasters, consisting of environmental disasters like hurricanes and wildfires; (ii) violence, consisting of episodes of violence such as public shootings and attacks conducted by humans.
We chose the top six U.S.-based events from 2019 to 2025 ranked by the number of fatalities \cite{gunviolencearchive, boulderfire,buffaloshooting,mauifire, lafires}. 
For each event, we specified an onset date, based on the date the event occurred, and a list of ten event keywords and phrases.
The events are listed in Table \ref{tab:events}.

\begin{table}[ht]
    \centering
    \begin{tabular}{l|c|c}
       \textbf{Event} & \textbf{Onset date} & \textbf{Category} \\ \hline
         Virginia Beach shooting & 2019-05-31 & Violence\\
         El Paso shooting & 2019-08-03 & Violence\\
         Buffalo shooting & 2022-05-14 & Violence \\
         Uvalde shooting & 2022-05-24 & Violence\\
         Monterey Park shooting & 2023-01-21 & Violence\\
         Boulder firebombing attack & 2025-05-30 & Violence\\
         Western Kentucky tornado & 2021-12-10 & Disaster \\
         Kansas wildfire outbreak & 2021-12-15 & Disaster \\
         Maui wildfires & 2023-08-08 & Disaster\\
         Hurricane Milton landfall & 2024-10-09 & Disaster \\
         Southern California fires & 2025-01-08 & Disaster \\
         Texas flash floods & 2025-07-04 & Disaster\\
         
    \end{tabular}
    \caption{List of events used in this study.}
    \label{tab:events}
\end{table}

We used the Global Database of Events (GDELT) \cite{leetaru2013gdelt} to retrieve news article URLs, resulting in coverage across 4,161 unique domains. We collected news articles spanning seven days prior to the event onset date and thirty days following the event, restricting results to the geographical location of the event. 
Then, we extracted the publication date, title and the first paragraph of articles using Goose3 article extraction framework for Python \cite{goose3}. 
All text was preprocessed with lemmatization, stop word removal, and near duplicates were dropped if their tf-idf similarity was above 0.9. 
Thus, for each event $i$, we produced a set $S(i)$ of documents. Documents were encoded using SentenceTransformers' \texttt{all-MiniLM-L6-v2} model \cite{reimers-2019-sentence-bert}, producing embedding vectors for each article.

We separated $S(i)$ into an \textbf{event subset} $S_e(i)$ and a \textbf{baseline subset} $S_b(i)$. Specifically, we assigned a document $d$ to $S_e(i)$ if its text contained two or more keywords from event $i$'s keyword set, otherwise, we assigned $d$ to $S_b(i)$. We refined the event set assignment by finding the ten nearest neighbors of each document in the baseline set $S_b(i)$ and moved them to the event set $S_e(i)$ if six or more of their neighbors were in $S_e(i)$ (majority voting). This refinement ensures that documents that are semantically similar to the event were included in the event set even if they do not contain any of the event's keywords.

\section{Creating Time Series of News Publications}

\paragraph{Time series extraction} We used daily time steps and subtracted the onset date from the publication date of each article. Thus, every event was represented in the time interval $[-7,+30]$ days, with time $t=0$ denoting the event onset date.

\paragraph{Document volume} The document volume was calculated as the proportion of documents published on each day to the total number of documents in the set. We expect the document volume of the event documents to exhibit a large peak early on, especially in the case of unpredictable events, as these events tend to dominate the news. The volume of the baseline is expected to follow the standard news cycle pattern: a slight volume increase midweek and slight decline in weekends.

\paragraph{Semantic drift} To measure the degree in which the framing of an event evolves, we computed the semantic drift by tracking how the centroid of the document embeddings move on a daily basis. The centroid $c_t$ is computed as the mean vector of all documents from time $t$: $c_t=\frac{1}{n_t}\sum_{d}\mathtt{x}_d$, where $\mathtt{x_d}$ is the embedding vector of document $d$ and $n_t$ is the number of documents on day $t$. We apply an exponential moving average to smoothen centroids and prevent abrupt displacement on days with few documents. The \textit{semantic drift} is the cosine distance between the centroid vectors on days $t$ and $t-1$.

\paragraph{Semantic dispersion} We measured the variation in the framing (dispersion) as the \textit{variance} of the distances between the daily centroid vector and the document vectors on each day. This signal describes how aligned or dispersed the discussion around an event is. 
We hypothesized that as the event unfolds, the discussion is dispersed and, over time, the framing is focused on a few points, becoming narrower, hence, dispersion drops.

\paragraph{Word relevance} To identify the dominant words at each time step, we computed the term frequency-inverse document frequency scores for the 300 most frequent words each day.

The above signals were calculated for each event in our dataset, and then were aggregated by the average value in each event category (Disaster, Violence) along with the standard error of the mean.

\section{Results and Discussion}

\subsection{Temporal Dynamics Analysis}
\begin{figure*}[ht]
    \centering
    \begin{subfigure}{0.33\linewidth}
        \includegraphics[width=\textwidth]{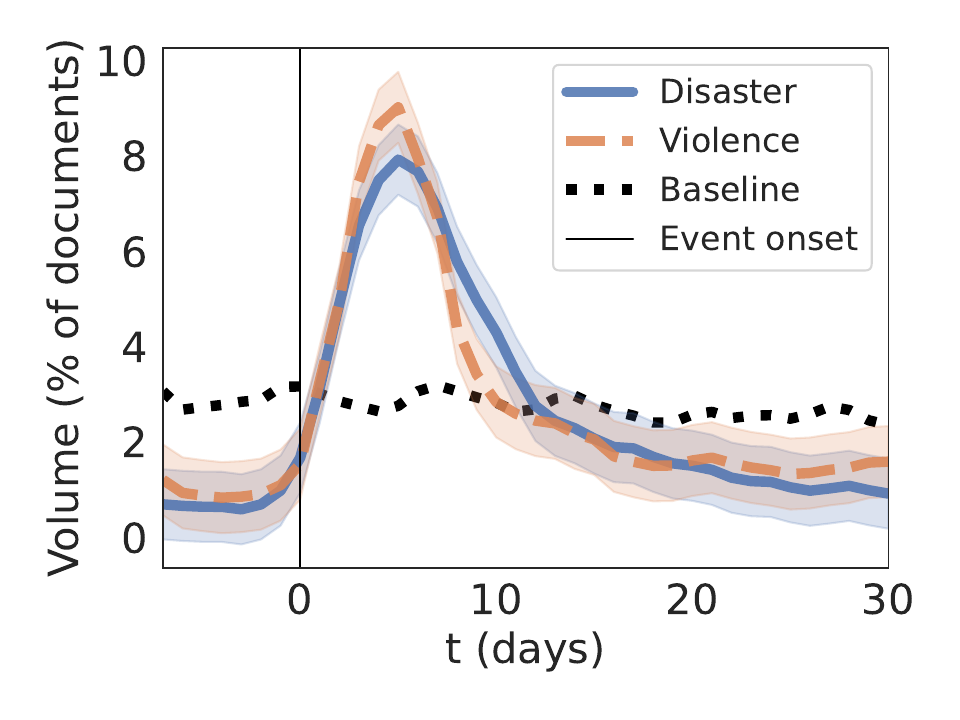}
        \caption{}
        \label{fig:volume}
    \end{subfigure}
    \begin{subfigure}{0.33\linewidth}
        \includegraphics[width=\textwidth]{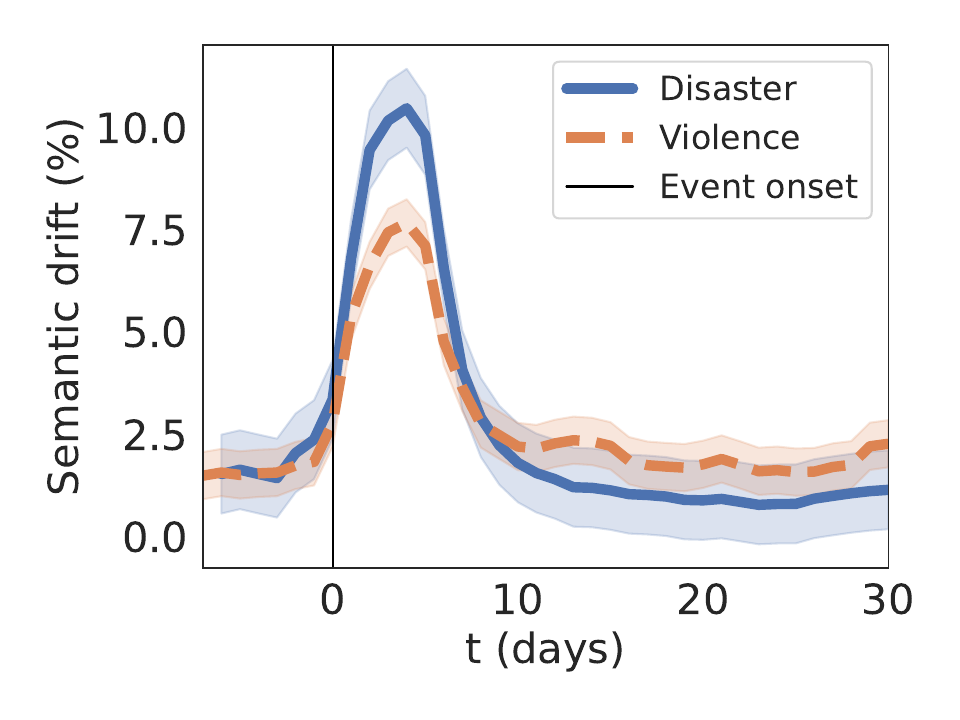}
        \caption{}
        \label{fig:drift}
    \end{subfigure}
    \begin{subfigure}{0.33\linewidth}
        \includegraphics[width=\textwidth]{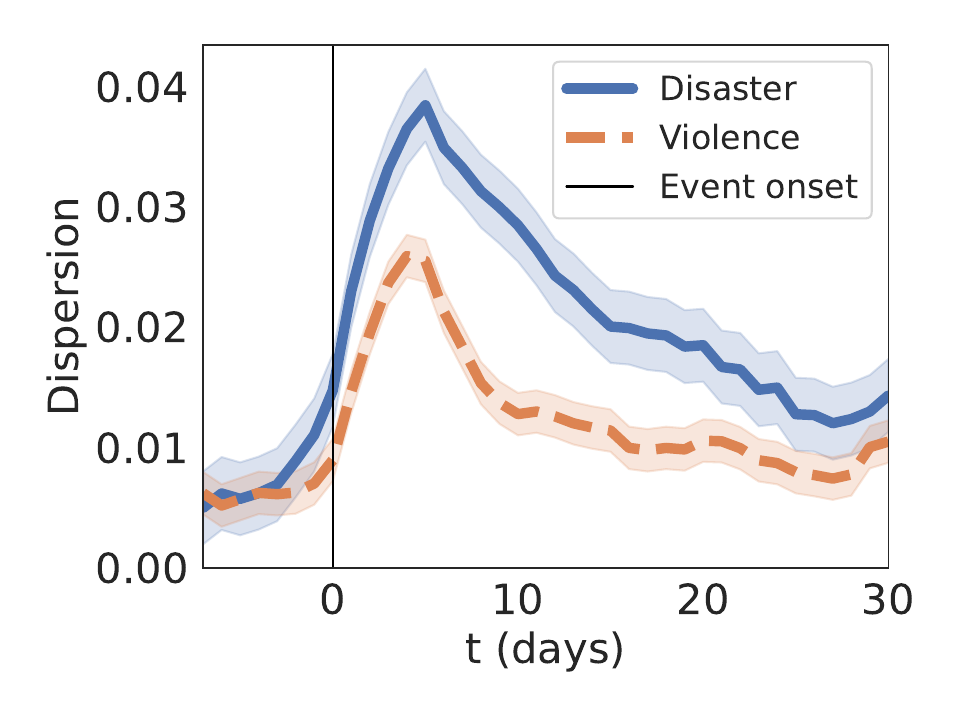}
        \caption{}
        \label{fig:dispersion}
    \end{subfigure}

    \caption{Time series of (a) volume, peaks at $t=5$; (b) semantic drift, peaks at $t=4$; (c) semantic dispersion, peaks at $t=5$. The baseline in (a) is the volume of articles published, the baseline for (b) and (c) is zero as we calculate the drift and dispersion with respect to event documents only. Shaded regions indicate 95\% CI.}
    \label{fig:timeseries}

\end{figure*}

\paragraph{Addressing RQ1} We measure media coverage with the publication volume, measured as the percentage of articles published within the time frame, peaks at $t=5$ days for both disaster and violence categories. Figure \ref{fig:volume} shows the average article volume in each category. At the peak, the amplitude of the volume curve is approximately $8\%\pm0.37$\%  for disaster and $9\%\pm0.37$\%  for violence up from a baseline volume of 3\%, this suggests that \textbf{violent events exhibit a slightly stronger early burst in volume than disasters}. The volume returns to baseline levels after 10 days.

    

The temporal dynamics of semantic drift also exhibit an early burst. 
The daily centroid drift used to measure the evolution of discussion shows early peaks for both event categories. Figure \ref{fig:drift} shows the time series of the semantic drift (in percentages of the total drift). 
However, \textbf{a stronger burst is observed in the framing drift in the disaster category}. 
This is likely due to the fact that disasters typically last longer than violent attacks--hence news stories may show a higher drift from one day to the next. Disaster and violence events peak at $t=5$ with amplitudes of $10\%\pm0.10\%$ and $7.5\%\pm0.18\%$ within 95\% CI, respectively.

The semantic dispersion follows the early burst trend, showing peaks at $t=4$ with amplitudes of $0.04\pm0.002$ and $0.025\pm0.001$ within 95\% CI for disaster and violence. Dispersion shows a more gradual decline after the peak, suggesting the initial reporting of events involves a broad framing, slowly converging to a narrow discussion. 
As seen in Figure \ref{fig:dispersion}, \textbf{disaster events exhibit a higher peak and a more gradual decline than violent events}. 
This is likely explained by the distinct nature of both types of events. Violent events are often unforeseen and sudden. 
The aftermath often includes fatalities, injuries and terror; followed by an investigation by authorities. Disaster events, on the other hand, may be predictable, but the consequences are always catastrophic and include fatalities, injuries, damage to properties and infrastructure, rescue and aid operations. The dispersion results underscore this distinction.

\subsection{Semantic Phase Analysis}

\begin{figure*}[ht]
    \centering
    \framebox{
    \begin{subfigure}{0.4\textwidth}
        \includegraphics[width=\textwidth]{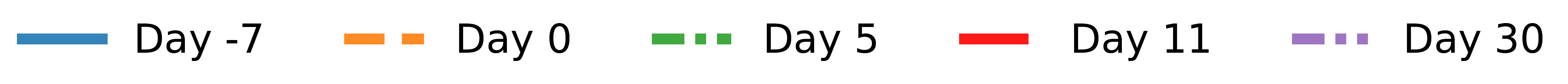}
    \end{subfigure}}\hspace{40pt}
    \fbox{\begin{subfigure}{0.4\textwidth}
        \includegraphics[width=\textwidth]{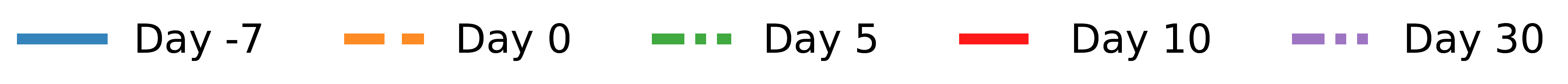}
    \end{subfigure}}
    
    \begin{subfigure}{0.24\linewidth}
        \includegraphics[width=\textwidth]{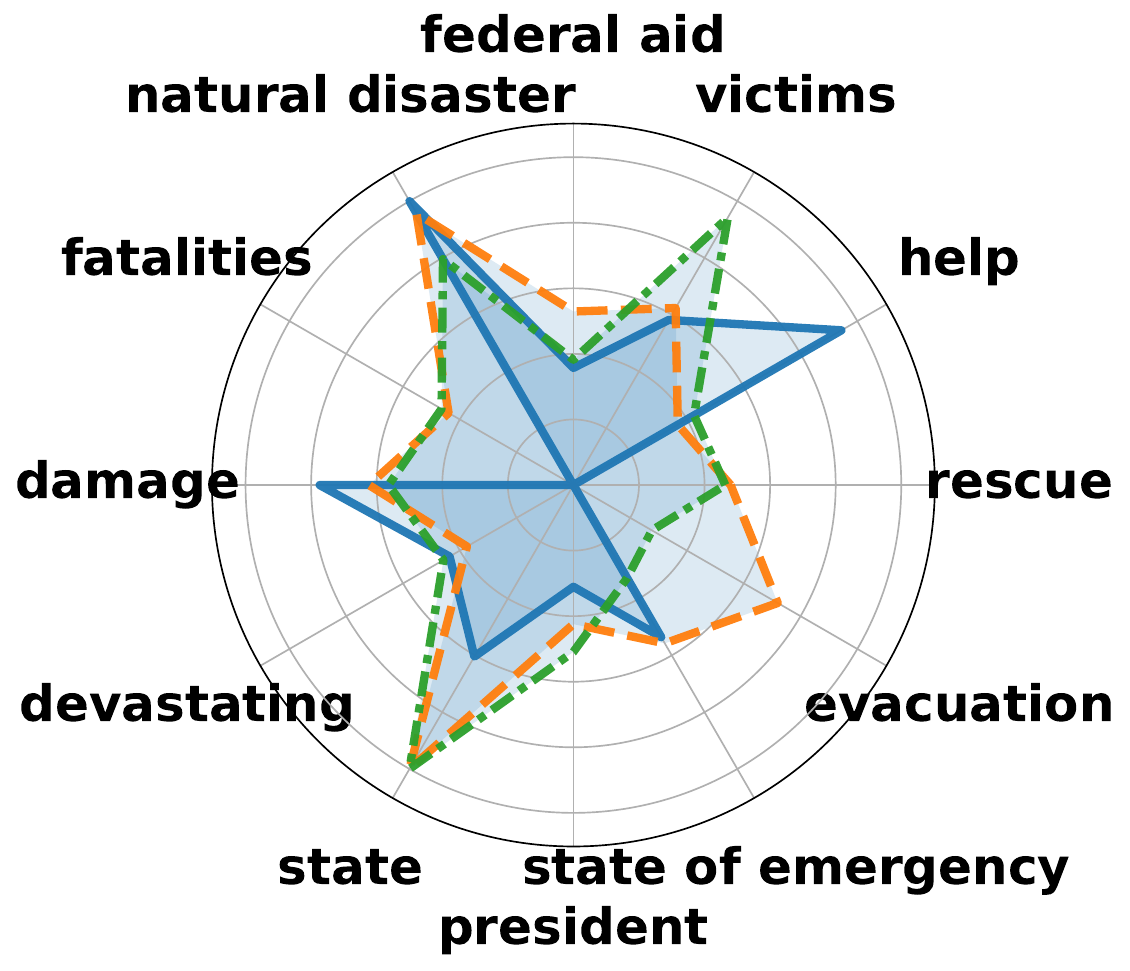}
        \caption{}
        \label{fig:disaster_phase1_2}
    \end{subfigure}
    \begin{subfigure}{0.24\linewidth}
                \includegraphics[width=\textwidth]{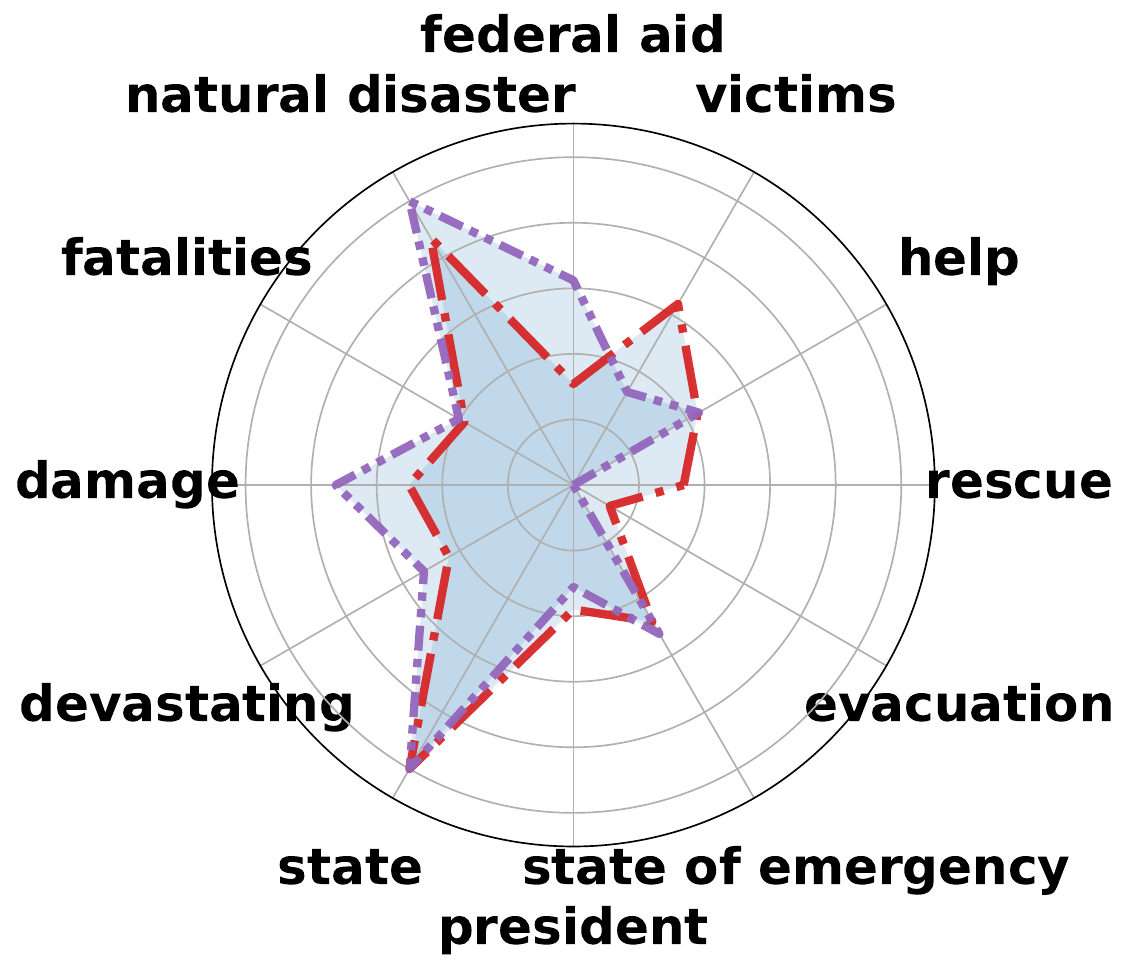}
        \caption{}
        \label{fig:disaster_phase3}
    \end{subfigure}
    \begin{subfigure}{0.24\linewidth}
                \includegraphics[width=\textwidth]{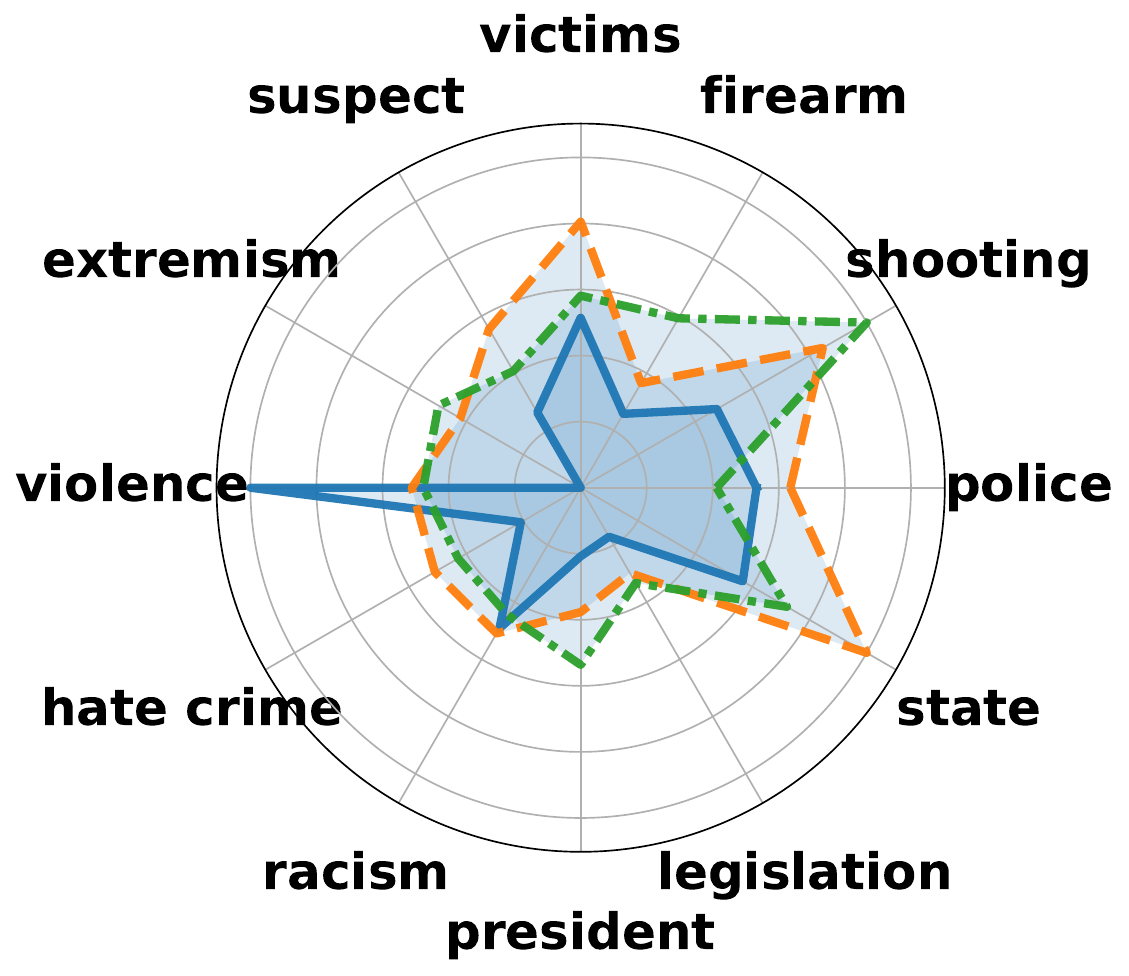}
        \caption{}
        \label{fig:violence_phase1_2}
    \end{subfigure}
    \begin{subfigure}{0.24\linewidth}
                \includegraphics[width=\textwidth]{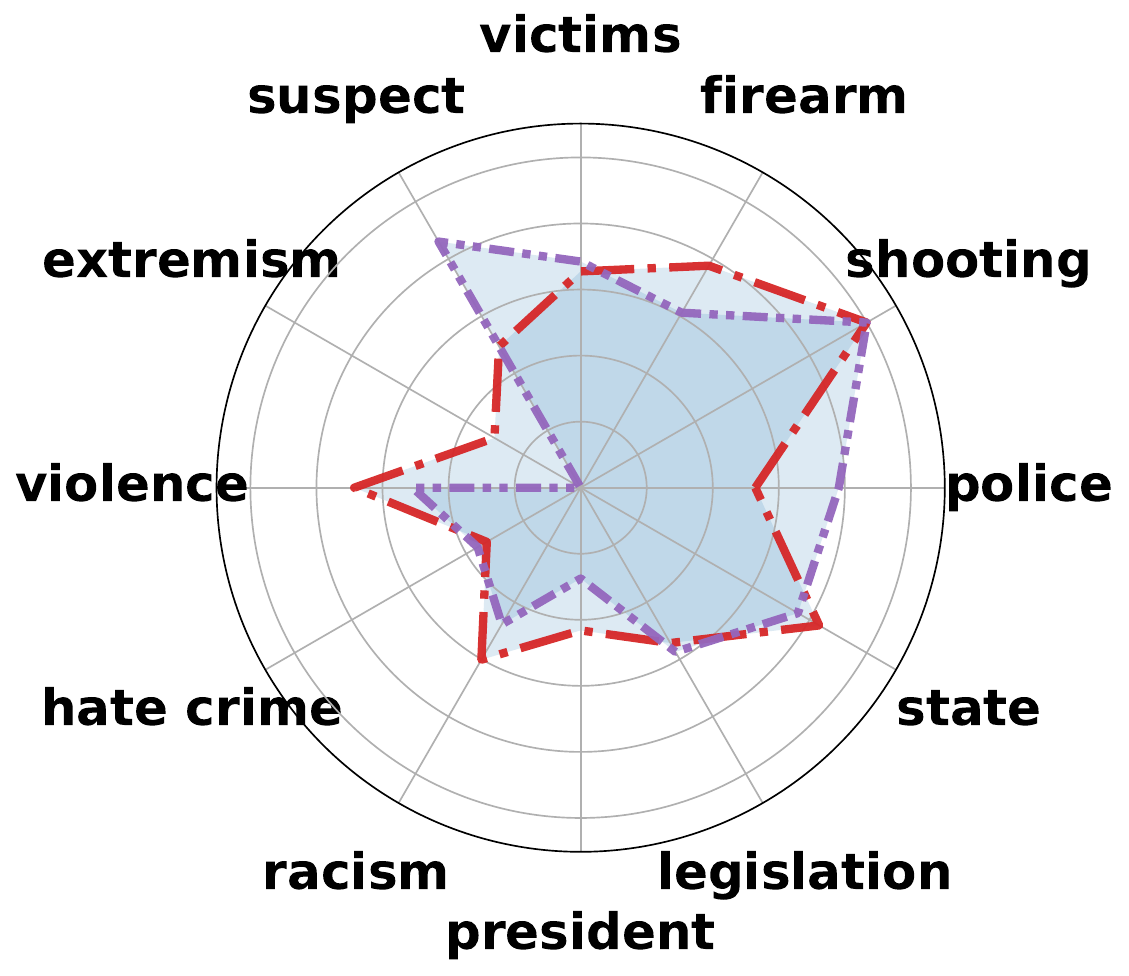}
        \caption{}
        \label{fig:violence_phase3}
    \end{subfigure}

    \caption{Radar charts showing the importance score of word groups across the semantic phases of the framing evolution. (a,b) disaster; (c,d) violence.}
    \label{fig:radar}

\end{figure*}

\paragraph{Addressing RQ2} To identify the factors that drive the semantic drift of events, we use the most relevant words, as measured by tf-idf, during the change points of the time series. Namely, we report the tf-idf scores of relevant words during the pre-event baseline ($t=-7$), volume peak and return to baseline. 

\paragraph{Disaster news cycle} The chart for words in the disaster category in Figure \ref{fig:radar} reveals a clear three-phase structure in news coverage. The first phase, spanning from day -7 to day 0 (Figure 2a), corresponds to the anticipatory stage of reporting. During this period, dominant terms such as “help”, “state”, and “natural disaster” indicate that coverage is primarily focused on preparedness, forecasting, and situational awareness. This phase represents a pre-impact information environment centered on prevention and readiness.

The second phase, occurring between day 0 and day 5 (Figure \ref{fig:disaster_phase1_2}), marks the crisis and impact stage of the news cycle. On the day of onset (day 0), coverage exhibits sharp increases in “evacuation”, “fatalities”, and “federal aid”, signaling a transition from forecast-based reporting to real-time emergency response. By day 5, semantic emphasis shifts toward “victims”, “help”, and continued reporting on fatalities, reflecting growing awareness of the human consequences of the disaster. This phase represents the peak of public attention and emotional intensity, with reporting centered on loss, disruption, and urgent intervention.

The third phase, extending from day 11 to day 30 (Figure \ref{fig:disaster_phase3}), captures the aftermath and recovery stage of coverage. By day 11, mentions of “federal aid” and “state of emergency” increase substantially, indicating the institutionalization of disaster response through policy action and formal declarations. By day 30, references to “fatalities”, “evacuation”, and “help” decline, while attention to “federal aid” continues to rise, signaling a shift of reporting from immediate crisis management toward long-term recovery and reconstruction.

\paragraph{Violence news cycle} The violence chart in Figure \ref{fig:radar} reveals an evolving pattern of media coverage following sudden violent events, illustrating how framing shifts from immediate incident reporting to broader political, social, and ideological discourse. The semantic evolution of violence coverage follows a progression from event description to policy debate and long-term ideological framing.

The first phase, spanning from day -7 to day 0 (Figure \ref{fig:violence_phase1_2}), represents the baseline context of violence reporting. During this period, coverage is anchored by general crime-related language, with prominent emphasis on “violence", “police", “shooting", and “state". These terms reflect routine reporting on violent crime, law enforcement activity, and public safety, establishing the semantic baseline against which the event-driven framing shift becomes visible.

The second phase, which occurs between day 0 and day 5 (Figure \ref{fig:violence_phase1_2}), marks the onset and crisis stage of the violence news cycle. On day 0, coverage shows sharp increases in “shooting", “victims", “state", and “police", indicating a transition from general crime reporting to focused coverage of a specific violent incident. Media framing becomes dominated by real-time developments, suspect identification, law enforcement response, and casualties. By day 5, attention expands toward “firearm”, “racism”, and “extremism”, suggesting a re-framing beyond a singular criminal act and toward broader social or ideological interpretations. 

The third phase, extending from day 10 to day 30 (Figure \ref{fig:violence_phase3}), reflects the aftermath of violence coverage. By day 10, “hate crime” emerges as a dominant term, accompanied by rising emphasis on “firearm”, “racism”, and “violence”. This indicates a media shift toward motive attribution and sociopolitical consequences. By day 30, “legislation”, “suspect”, and “shooting” become prominent, reflecting the transition from incident reporting to policy debate and legal proceedings. Media discourse during this phase centers on gun control legislation, criminal prosecution, and broader discussions of systemic violence and extremism.

The semantic evolution of disaster and violence coverage reveals two distinct but structurally similar news-cycle patterns. In both cases, media attention surges after event onset, peaks during the crisis phase, and gradually declines as coverage shifts toward long-term consequences. However, disaster reporting is primarily operational and recovery-focused, while violence reporting becomes increasingly politicized and ideologically framed over time.

\section{Limitations and Ethical Considerations}

This study was conducted on a set of articles collected from hundreds of websites. 
It includes reports made not exclusively by the mainstream media but also local and alternative media. 
Alternative media has played an increasingly important role in news reporting in recent years, hence, we believe it is important to include these sources to better reflect the online news reporting landscape.
As a result, inaccurate, false, and biased information may be present in the data.

We describe the dynamics of news reporting of events whose consequences include fatalities and injuries to individuals. These events are often at the center of disinformation campaigns  and conspiracy theories. However, we do not anticipate that the results and findings we report here can cause negative societal impacts.

\section{Conclusion and Future Work}

We presented an analysis of temporal dynamics of the news coverage of violent and catastrophic events. 
Our method analyzes how stories develop by examining a time frame of 7 days prior to the event up to 30 days after the event onset using document volume, semantic drift and semantic dispersion to measure the framing evolution.
Results show that volume, drift and dispersion peak a few days after the onset. However, disaster events show a higher amplitude, suggesting a stronger drift and breadth of framing as the events unfold. By analyzing the importance of common words, we characterize the drivers of framing evolution as words that pull the narrative towards discussions in the event's aftermath, like evacuation and assistance for disasters, and legislation and extremism for violence.

We plan to further extend this analysis to other events and categories, such as social issues, economic and political crises. Furthermore, we plan to extend the semantic phase analysis to investigate the influence of rumors, misinformation and disinformation on framing evolution.

\clearpage
\bibliography{ref}

\input{checklist}

\end{document}

%% file: checklist.tex
\newcommand{\answerTODO}[1]{\textbf{#1}}
\section*{Paper Checklist}

\begin{enumerate}
    \item  Would answering this research question advance science without violating social contracts, such as violating privacy norms, perpetuating unfair profiling, exacerbating the socio-economic divide, or implying disrespect to societies or cultures?
    \answerTODO{Yes}
  \item Do your main claims in the abstract and introduction accurately reflect the paper's contributions and scope?
    \answerTODO{Yes}
   \item Do you clarify how the proposed methodological approach is appropriate for the claims made? 
    \answerTODO{Yes}
   \item Do you clarify what are possible artifacts in the data used, given population-specific distributions?
    \answerTODO{N/A}
  \item Did you describe the limitations of your work?
    \answerTODO{Yes, limitations are discussed in the “Limitations and Ethical Considerations” section.}
  \item Did you discuss any potential negative societal impacts of your work?
    \answerTODO{Yes, in the “Limitations and Ethical Considerations” section.}
      \item Did you discuss any potential misuse of your work?
    \answerTODO{N/A}
    \item Did you describe steps taken to prevent or mitigate potential negative outcomes of the research, such as data and model documentation, data anonymization, responsible release, access control, and the reproducibility of findings?
    \answerTODO{Yes}
  \item Have you read the ethics review guidelines and ensured that your paper conforms to them?
    \answerTODO{Yes}
\end{enumerate}

Additionally, if your study involves hypotheses testing...
\begin{enumerate}
  \item Did you clearly state the assumptions underlying all theoretical results?
    \answerTODO{N/A}
  \item Have you provided justifications for all theoretical results?
    \answerTODO{N/A}
  \item Did you discuss competing hypotheses or theories that might challenge or complement your theoretical results?
    \answerTODO{N/A}
  \item Have you considered alternative mechanisms or explanations that might account for the same outcomes observed in your study?
    \answerTODO{N/A}
  \item Did you address potential biases or limitations in your theoretical framework?
    \answerTODO{N/A}
  \item Have you related your theoretical results to the existing literature in social science?
    \answerTODO{N/A}
  \item Did you discuss the implications of your theoretical results for policy, practice, or further research in the social science domain?
    \answerTODO{N/A}
\end{enumerate}

Additionally, if you are including theoretical proofs...
\begin{enumerate}
  \item Did you state the full set of assumptions of all theoretical results?
    \answerTODO{N/A}
	\item Did you include complete proofs of all theoretical results?
    \answerTODO{N/A}
\end{enumerate}

Additionally, if you ran machine learning experiments...
\begin{enumerate}
  \item Did you include the code, data, and instructions needed to reproduce the main experimental results (either in the supplemental material or as a URL)?
    \answerTODO{Yes, in supplemental material.}
  \item Did you specify all the training details (e.g., data splits, hyperparameters, how they were chosen)?
    \answerTODO{N/A, we did not train a model.}
     \item Did you report error bars (e.g., with respect to the random seed after running experiments multiple times)?
    \answerTODO{Yes, error bars are shown in figures.}
	\item Did you include the total amount of compute and the type of resources used (e.g., type of GPUs, internal cluster, or cloud provider)?
    \answerTODO{No, we did not use any external computer resources.}
     \item Do you justify how the proposed evaluation is sufficient and appropriate to the claims made? 
    \answerTODO{Yes}
     \item Do you discuss what is ``the cost`` of misclassification and fault (in)tolerance?
    \answerTODO{N/A}
  
\end{enumerate}

Additionally, if you are using existing assets (e.g., code, data, models) or curating/releasing new assets, \textbf{without compromising anonymity}...
\begin{enumerate}
  \item If your work uses existing assets, did you cite the creators?
    \answerTODO{Yes, GDELT database was referenced.}
  \item Did you mention the license of the assets?
    \answerTODO{N/A}
  \item Did you include any new assets in the supplemental material or as a URL?
    \answerTODO{N/A}
  \item Did you discuss whether and how consent was obtained from people whose data you're using/curating?
    \answerTODO{N/A}
  \item Did you discuss whether the data you are using/curating contains personally identifiable information or offensive content?
    \answerTODO{N/A}
\item If you are curating or releasing new datasets, did you discuss how you intend to make your datasets FAIR?
\answerTODO{N/A}
\item If you are curating or releasing new datasets, did you create a Datasheet for the Dataset? 
\answerTODO{N/A}
\end{enumerate}

Additionally, if you used crowdsourcing or conducted research with human subjects, \textbf{without compromising anonymity}...
\begin{enumerate}
  \item Did you include the full text of instructions given to participants and screenshots?
    \answerTODO{N/A}
  \item Did you describe any potential participant risks, with mentions of Institutional Review Board (IRB) approvals?
    \answerTODO{N/A}
  \item Did you include the estimated hourly wage paid to participants and the total amount spent on participant compensation?
    \answerTODO{N/A}
   \item Did you discuss how data is stored, shared, and deidentified?
   \answerTODO{N/A}
\end{enumerate}

%% file: ref.bib
@inproceedings{leetaru2013gdelt,
  title={GDELT: Global Data on Events, Location, and Tone, 1979--2012},
  author={Leetaru, Kalev and Schrodt, Philip A.},
  booktitle={ISA Annual Convention},
  year={2013},
  url={https://www.gdeltproject.org}
}

@misc{gunviolencearchive,
  title={Gun Violence Archive},
  author={{Gun Violence Archive}},
  year={2025},
  note={\url{https://www.gunviolencearchive.org/}. Accessed 2026-01-15},
  url={https://www.gunviolencearchive.org/}
}

@misc{mauifire,
    title={Maui wildfire one of deadliest in U.S. history},
    author={NFPA},
    year={2023},
    note={\url{https://www.nfpa.org/news-blogs-and-articles/blogs/2023/09/19/maui-wildfire-one-of-deadliest-in-us-history}. Accessed 2026-01-15},
    url={https://www.nfpa.org/news-blogs-and-articles/blogs/2023/09/19/maui-wildfire-one-of-deadliest-in-us-history}
}

@misc{lafires,
    title={2025 Los Angeles Fires},
    author={CA.gov},
    year={2025},
    note={\url{https://www.ca.gov/LAfires/}. Accessed 2026-01-15},
    url={https://www.ca.gov/LAfires/}
}

@misc{boulderfire,
    title={Alleged Perpetrator of Terror Attack in Colorado Charged with Hate Crimes},
    author={Justice.gov},
    year={2025},
    note={\url{https://www.justice.gov/opa/pr/alleged-perpetrator-terror-attack-colorado-charged-hate-crimes}. Accessed 2026-01-15},
    url={https://www.justice.gov/opa/pr/alleged-perpetrator-terror-attack-colorado-charged-hate-crimes}
}

@misc{buffaloshooting,
    title={Man who killed 10 in Buffalo mass shooting wants death penalty trial relocated to NYC},
    author={{USA Today}},
    year={2025},
    note={\url{https://www.usatoday.com/story/news/nation/2025/04/02/buffalo-racist-massacre-death-penalty/82781659007/}. Accessed 2026-01-15},
    url={https://www.usatoday.com/story/news/nation/2025/04/02/buffalo-racist-massacre-death-penalty/82781659007/}
}

@misc{goose3,
    title={Goose3 Article Extractor},
    author={Mahmoud Lababidi},
    year={2018},
    note={\url{https://github.com/goose3/goose3}. Accessed 2026-01-13},
    url={https://github.com/goose3/goose3}
}

@inproceedings{reimers-2019-sentence-bert,
  title = "Sentence-BERT: Sentence Embeddings using Siamese BERT-Networks",
  author = "Reimers, Nils and Gurevych, Iryna",
  booktitle = "Proceedings of the 2019 Conference on Empirical Methods in Natural Language Processing",
  month = "11",
  year = "2019",
  publisher = "Association for Computational Linguistics",
  url = "https://arxiv.org/abs/1908.10084",
}

@article{silva2021news,
  title={The news media's framing of mass shootings: Gun access, mental illness, violent entertainment, and terrorism},
  author={Silva, Jason R},
  journal={Russ. J. Econ. \& L.},
  pages={332},
  year={2021},
  publisher={HeinOnline}
}

@article{park2025applying,
  title={Applying semantic network analysis to explore the relationship between media ideology and editorial coverage of COVID-19: S. Park et al.},
  author={Park, Sejung and Sridharan, Nisha and Kwon, K Hazel},
  journal={Quality \& Quantity},
  volume={59},
  number={Suppl 2},
  pages={1283--1303},
  year={2025},
  publisher={Springer}
}

@article{geiss2025inflation,
  title={Inflation of crisis coverage? Tracking and explaining the changes in crisis labeling and crisis news wave salience 1785--2020},
  author={Gei{\ss}, Stefan and Viehmann, Christina and Kelly, Conor A},
  journal={Journal of Communication},
  volume={75},
  number={1},
  pages={27--41},
  year={2025},
  publisher={Oxford University Press}
}

@inproceedings{an2017gets,
  title={What gets media attention and how media attention evolves over time: large-scale empirical evidence from 196 countries},
  author={An, Jisun and Kwak, Haewon},
  booktitle={Proceedings of the International AAAI Conference on Web and Social Media},
  volume={11},
  number={1},
  pages={464--467},
  year={2017}
}

@article{wormwood2019psychological,
  title={Psychological impact of mass violence depends on affective tone of media content},
  author={Wormwood, Jolie Baumann and Lin, Yu-Ru and Lynn, Spencer K and Barrett, Lisa Feldman and Quigley, Karen S},
  journal={PLoS one},
  volume={14},
  number={4},
  pages={e0213891},
  year={2019},
  publisher={Public Library of Science San Francisco, CA USA}
}

@article{tierney2006metaphors,
  title={Metaphors matter: Disaster myths, media frames, and their consequences in Hurricane Katrina},
  author={Tierney, Kathleen and Bevc, Christine and Kuligowski, Erica},
  journal={The annals of the American academy of political and social science},
  volume={604},
  number={1},
  pages={57--81},
  year={2006},
  publisher={Sage Publications Sage CA: Thousand Oaks, CA}
}

@article{livingston2003gatekeeping,
  title={Gatekeeping, indexing, and live-event news: Is technology altering the construction of news?},
  author={Livingston, Steven and Bennett, W Lance},
  journal={Political Communication,},
  volume={20},
  number={4},
  pages={363--380},
  year={2003},
  publisher={Taylor \& Francis}
}

@article{nielsen2016challenges,
  title={Challenges and opportunities for news media and journalism in an increasingly digital, mobile, and social media environment},
  author={Nielsen, Rasmus and Cornia, Alessio and Kalogeropoulos, Antonis},
  journal={Reuters Institute for the Study of Journalism},
  year={2016},
  publisher={Reuters Institute for the Study of Journalism}
}
